\documentclass[runningheads]{llncs}
\usepackage{graphicx}
\usepackage[utf8]{inputenc} 
\usepackage[T1]{fontenc}    
\usepackage{hyperref}       
\usepackage{url}            
\usepackage{booktabs}       
\usepackage{amsfonts}       
\usepackage{nicefrac}       
\usepackage{microtype}      
\usepackage{amssymb}
\usepackage{array}
\usepackage{graphicx}
\usepackage{subcaption}
\usepackage{booktabs}
\usepackage{url}
\usepackage{algorithm}
\usepackage{algpseudocode}
\usepackage{mathtools}
\usepackage{amsmath}
\usepackage{blindtext}
\usepackage{enumitem}
\usepackage{xcolor}
\usepackage{amsbsy}
\usepackage{bm}
\begin{document}
	\title{Efficient Interpretation of Deep Learning Models Using Graph Structure and Cooperative Game Theory: Application to ASD Biomarker Discovery }
	\titlerunning{Abbreviated paper title}
	%
	\author{Xiaoxiao Li\inst{1} \and
	Nicha C. Dvornek\inst{2} \and
	Yuan Zhou\inst{2}\and
	Juntang Zhuang \inst{1}\and \\
	Pamela Ventola \inst{3}\and
	James S. Duncan \inst{1,2,4,5}}
	%
	%
    \institute{Biomedical Engineering, Yale University, New Haven, CT USA \and
	Radiology \& Biomedical Imaging, Yale School of Medicine, New Haven, CT USA \and
	Child Study Center, Yale School of Medicine, New Haven, CT USA \and
	Electrical Engineering, Yale University, New Haven, CT, USA \and
    Statistics \& Data Science, Yale University New Haven, CT, USA
	}
	%
	\maketitle              
\vspace{-7mm}
\begin{abstract}
     Discovering imaging biomarkers for autism spectrum disorder (ASD) is critical to help explain ASD and predict or monitor treatment outcomes. Toward this end, deep learning classifiers have recently been used for identifying ASD from functional magnetic resonance imaging (fMRI) with higher accuracy than traditional learning strategies. However, a key challenge with deep learning models is understanding just what image features the network is using, which can in turn be used to define the biomarkers. Current methods extract biomarkers, i.e., important features, by looking at how the prediction changes if "ignoring" one feature at a time. However, this can lead to serious errors if the features are conditionally dependent. In this work, we go beyond looking at only individual features by using Shapley value explanation (SVE) from cooperative game theory. Cooperative game theory is advantageous here because it directly considers the interaction between features and can be applied to any machine learning method, making it a novel, more accurate way of determining instance-wise biomarker importance from deep learning models. A barrier to using SVE is its computational complexity: $2^N$ given $N$ features. We explicitly reduce the complexity of SVE computation by two approaches based on the underlying graph structure of the input data: 1) only consider the centralized coalition of each feature; 2) a hierarchical pipeline which first clusters features into small communities, then applies SVE in each community. Monte Carlo approximation can be used for large permutation sets. We first validate our methods on the MNIST dataset and compare to human perception. Next, to insure plausibility of our biomarker results, we train a Random Forest (RF) to classify ASD/control subjects from fMRI and compare SVE results to standard RF-based feature importance. Finally, we show initial results on ranked fMRI biomarkers using SVE on a deep learning classifier for the ASD/control dataset.
\end{abstract}
\vspace{-5mm}
	\section{Introduction}
	Autism spectrum disorder (ASD) affects the structure and function of the brain. To better target the underlying roots of ASD for diagnosis and treatment, efforts to identify reliable biomarkers are growing \cite{goldani2014biomarkers}. Deep learning models have been used in fMRI analysis \cite{li2018brain}, which is used to characterize the brain changes that occur in ASD \cite{Kaiser07122010}. However, how the different brain regions coordinate on the deep convolutional neural network (DNN) has not been previously explored. When features are not independent, Shapley value explanation (SVE) is a useful tool to study each feature's contribution \cite{lundberg2017unified,chen2018shapley,kononenko2010efficient}.  The methods are based on fundamental concepts from cooperative game theory \cite{shapley1953value}, which assigns a unique distribution (among the players) of a total surplus generated by the coalition of all players in the cooperative game. However, if the interactive features' dimensions are high, SVE becomes computationally consuming (exponential time complexity). 
	
	The innovations of this study include: 1) We applied SVE on interactive features' prediction power analysis; 2) Our proposed method does not require retraining the classifier; 3) To handle the high dimensional inputs of the DNN classifier, we propose two methods to reduce the dimension of SVE testing features, once the underlying graph structure of features is defined; and 4) Different from kernel SHAP proposed in \cite{lundberg2017unified}, as a model interpreter, our proposed methods do not require model approximation. In section 2, we introduce the background on cooperative game theory.  In section 3, we propose the two approaches to approximate Shapley value. We also show the approximation is true under certain assumptions. Three experiments are given in section 4 to show the feasibility and advantage of our proposed methods. 
	\section{Background on Cooperative Game Theory}
	\subsection{Shapley Value}
	Our approach to analyzing the contributions of individual nodes to the overall network is the assignment of Shapley values. The Shapley value is a means of fairly portioning the collective profit attained by a coalition of players, based on the relative contributions of the players in some game.  Let $\mathcal{N}=\{1,2,\dots,N\}$ be the set of all the players, $S \subset \mathcal{N}$ be a subset of players forming a coalition within this game, and $v : 2^ \mathcal{N}\rightarrow{\mathbb{R}}$ be the function that assigns a real numbered profit to the subset $S$ of players. By definition, for any $v$, $v(\varnothing) = 0$, here $\varnothing$ is the empty set. A Shapley value is assigned by a Shapley function $\Phi: \mathcal{N} \rightarrow{\mathbb{R}}$, which associates each player in $\mathcal{N}$ with a real number and which is uniquely defined by the following axioms \cite{shapley1953value}: 1. \textit{Efficiency}; 2. \textit{Symmetry}; 3. \textit{Dummy}; and 4. \textit{Additivity}. In our context, we are interested in the brain regions that discriminate ASD and control subjects. Classification prediction score is the total value to be distributed, and each brain region is a player, which will be assigned a unique reward (i.e. importance score) by its contribution to the classifier.
	
	\subsection{Challenges Of Using Shapley Value}
\begin{figure}[t]
		\centering
		\includegraphics[width=11cm,height=2.5cm]{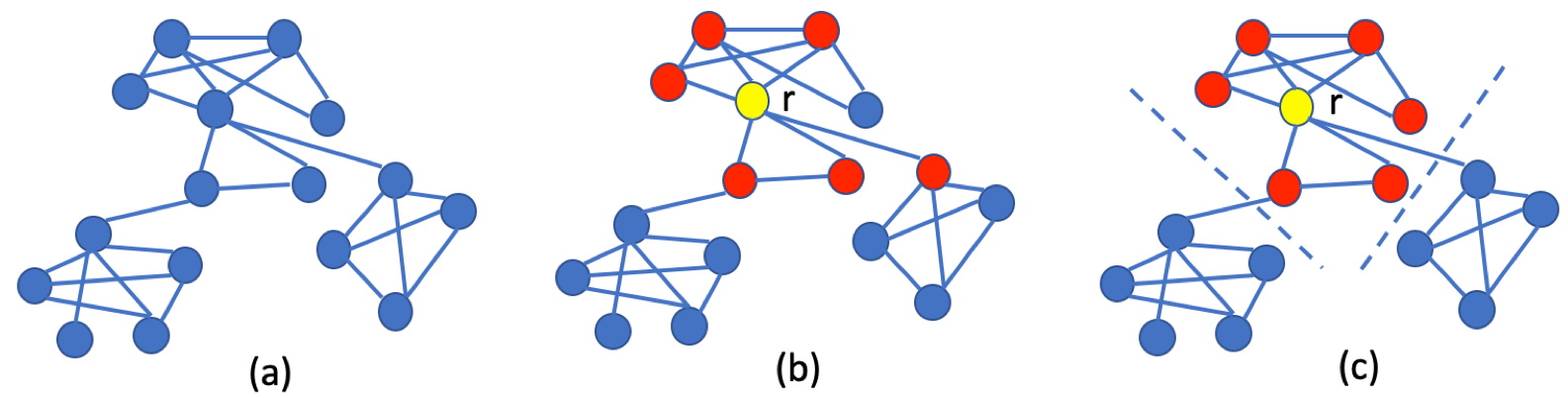}
		\caption{a) Toy visualization of graph structure of the input data. When estimating the contribution of feature $r$ (yellow), b) C-SVE considers $r$'s directly connected neighbors (red) and c) H-SVE considers the community (red) to which $r$ belongs.}
		\label{toynetwork}   
		\vspace{-4mm}
	\end{figure}

	While Shapley values give a more accurate interpretation of the importance of each player in a coalition, their calculation is expensive. When the number of features (i.e., players in the game) is a massive $N$, the computational complexity is $2^N$, which is especially expensive if the model is slow to run.	
	We propose addressing this computational challenge by utilizing the graph structure of the data. Consider the case when the underlying graph structure of the data is sparsely connected, e.g., the sparse brain functional network. Under this observation, we propose two approaches (Fig. \ref{toynetwork}) to simplify Shapley value calculation. \textit{Method I} only considers the centralized coalition of each player to reduce the number of permutation cases by assigning weight 0 to features that rarely collaborate. \textit{Method II} first applies community detection on the feature connectivity network to cluster similar features (forming different games and teams), then within the selected communities, assigns a feature's contribution by SVE. 
	
	\section{Methods}
	\vspace{-1mm}
	In classification tasks, only certain features in a given input provide evidence for the classification decision. For a given prediction, the classifier assigns a relevance value to each input feature with respect to a class label $Y\in \mathcal{C}$. The probability of class $Y$ for input $\bm{X}= ( X_1, X_2, \dots, X_N) $ is given by the predictive score of the DNN model $f: \mathcal{D}\rightarrow \mathbb{R}^{|\mathcal{C}|}$ where $\mathcal{D}$ is the domain for input $X$ and each component of the output of $f$ represents the conditional probability of assigning a class label, i.e.  $  p(Y|\bm{X})$. 
	
	The basic idea used in prediction difference analysis \cite{zintgraf2017visualizing} is that the relevance of a feature $x_i$ can be estimated by measuring how the prediction changes if the feature is unknown. Here we extend this setting by considering the interaction of a set of different features instead of examining the features one by one. Denote the image corrupted at a feature set $S \subseteq \mathcal{N}$ as $\bm{X}_{\mathcal{N}\setminus S}$.  To calculate $p(Y|\bm{X}_{\mathcal{N}\setminus S})$, following \cite{zintgraf2017visualizing}, we marginalize out the corrupted feature set $S$:
	\begin{equation}
	\label{margin}
	p(Y|\bm{X}_{\mathcal{N}\setminus S}) = \mathbb{E}_{\bm{X}_{S}\sim p(\bm{X}_{S}|\bm{X}_{\mathcal{N}\setminus S})}p(Y|\bm{X}_{\mathcal{N}\setminus S},\bm{X}_{S}).
	\end{equation}
	Denote $v_{\bm{X}}$ the importance score evaluation function for input $\bm{X}$. The prediction power for the $r$th feature is the weighted sum of all possible marginal contributions:\\
	\begin{equation}
	\label{origshap}
	\Phi_r(v_{\bm{X}})  = \frac{1}{ |\mathcal{N}|}\sum_{S \subseteq \mathcal{N}\setminus\{r\}} \left ( \begin{array}{c}
	 |\mathcal{N}|- 1 \\
	|S| 
	\end{array}
	\right ) ^ {-1}  (v_{\bm{X}}(S\cup\{r\})-v_{\bm{X}}(S)).
	\end{equation}
	Similar to \cite{chen2018shapley}, we introduce the \textit{importance score} of a feature set $S$
	\begin{equation}
	v_{\bm{X}}(S)\coloneqq \mathbb{E}_Y[-\log\frac{1}{p(Y|\bm{X})}|\bm{X}] - \mathbb{E}_Y[-\log\frac{1}{p(Y|\bm{X}_{\mathcal{N}\setminus S})}|\bm{X}],
	\end{equation}
which can be interpreted as the negative of the expected number of bits required to encode the output of the model based on the input $\bm{X}_{\mathcal{N}\setminus S}$. \\
	\textbf{Theorem 1} $\langle\mathcal{N} = \{1,2,\dots,N\},v_{\bm{X}}\rangle$ \textit{ is a cooperative form game and } $\Phi(v_{\bm{X}}) = (\Phi_1, \Phi_2,\dots,\Phi_N)$ \textit{corresponds to the game's Shapley value}.\\
	The proof can be directly borrowed from \cite{kononenko2010efficient} showing it has a unique solution and  satisfies Axioms 1-4.  

An illustrative example is Boolean logic expression, $\text{OR}((x_1,x_2))=1$ when $x_1$ or $x_2$ is one  and zero otherwise for $\mathcal{N}=\{1,2\}$, $\mathcal{D} =\{0,1\} \times \{0,1\}$. Suppose $p(Y=1|\bm{X})=\text{OR}(\bm{X})$ and the base of the logarithm is 2. We aim to find the contributions of predicting 1 given input $\bm{X}=(1,1)$.  If both values of $\bm{X}$ are unknown, one can predict that the probability of the result being 1 is $\frac{3}{4}$. We have $v_{\bm{X}}(\varnothing)= 0 - (-\log\frac{1}{1}) = 0$, $v_{\bm{X}}(\{1\}) = v_{\bm{X}}(\{2\}) = 0 - (-\log(\frac{1}{1})) = 0$ and total value $v_{\bm{X}}(\{1,2\}) = 0 - (-\log\frac{4}{3}) = \log\frac{4}{3}$. Therefore the contributions of each feature are: $\Phi_1 = \frac{1}{2}[(v_{\bm{X}}(\{1\}) - v_{\bm{X}}(\varnothing)) + (v_{\bm{X}}(\{1,2\})- v_{\bm{X}}(\{2\}))]= \frac{1}{2}[(0-0)+(\log\frac{4}{3}-0)]=\frac{1}{2}\log\frac{4}{3}$ and $\Phi_2 = \frac{1}{2}[(v_{\bm{X}}(\{2\}) - v_{\bm{X}}(\varnothing)) + (v_{\bm{X}}(\{1,2\})- v_{\bm{X}}(\{1\}))]= \frac{1}{2}[(0-0)+(\log\frac{4}{3}-0)]=\frac{1}{2}\log\frac{4}{3}$. The generated contributions reveal that both features contribute the same amount towards the prediction being 1 given input $(1,1)$. In addition, we can interpret there is coalition between the two players, since $v_{\bm{X}}(\{1,2\}) > v_{\bm{X}}(\{1\}) + v_{\bm{X}}(\{2\})$. However, we will get the myopic conclusion that both features are unimportant by only ignoring a single feature, because given one feature $X_i = 1$, $p = 1$ is for sure.


	With the underlying structure of data, we have prior knowledge that some features of the data set are barely connected; in other words, there is very likely no coalition between these features. We define a connected graph $\mathcal{G = (V,E)}$ with nodes $\mathcal{V}$ and edges $\mathcal{E}$. Given an adjacency matrix $\mathcal{A} = (a_{ij})$ of the undirected graph $\mathcal{G}$ (for example, the Pearson correlation of mean time series of brain regions), we use a threshold $th$ to binarize $a_{ij}$, i.e. $a_{ij}^b=1$ when $a_{ij}>th$ and zero otherwise, resulting in a sparsely connected graph.
	\subsection{Method I: Centralized Shapley Value Explanation (C-SVE)}
	\vspace{-1mm}
	For a given feature $i$, its 1-step connected \textit{neighborhood} is defined by the set $ \mathcal{N}_i\coloneqq \{j \in \mathcal{V}|	{a}_{ij}^{b} = 1 \}$. As an approximation, we propose Centralized Shapley Value Explanation (C-SVE), which only calculates the marginal contribution when a feature collaborates with its neighbors.\\
	\textbf{Definition 1}. \textit{Given classifier $f$ and sample $\bm{X}$, the C-SVE assigns the prediction power on feature $r$ by}
	\begin{equation}
	\label{csve}
	\hat{\Phi}^C_r(v_{\bm{X}})   = \frac{1}{|\mathcal{N}_r|}\sum_{S \subseteq \mathcal{N}_r\setminus\{r\}} \left ( \begin{array}{c}
	|\mathcal{N}_r| - 1 \\
	|S| 
	\end{array}
	\right ) ^ {-1} (v_{\bm{X}}(S\cup\{r\})-v_{\bm{X}}(S)).
    \vspace{-1mm}
	\end{equation}
	The coefficients in front of the marginal contributions is a weighted transformation of the original SVE form (in Eq. (\ref{origshap})), where instead of assigning each permutation the same weight, sets not belonging to the \textit{neighborhood} were assigned 0 weight. In practice, we can reject the non-coalition permutations  and average the Shapley values for the remaining terms. \\
	\textbf{Theorem 2} \textit{ We have $\hat{\Phi}^C_{\bm{X}}(r) = \Phi_{\bm{X}}(r)$ almost surely if we have $X_r \perp \bm{X}_{\mathcal{N}\setminus \mathcal{N}_r} | \bm{X}_U$ and $X_r \! \perp \ \bm{X}_{\mathcal{N}\setminus \mathcal{N}_r} | \bm{X}_{U}, Y$ for any $U \subset \mathcal{N}_r \setminus \{r\}$}.
	
	The proof is shown in Appendix A. It is important to show that our proposed approximation is a good one. We can easily check the necessary condition that for $k \notin \mathcal{N}_r$, the angle between the average time series $\bar{X}_r$ in ROI $r$ and $\bar{X}_k$ in ROI $k$  satisfies $\cos(\bar{X}_r,\bar{X}_k)< \epsilon$, which corresponds to the small edge weight $(\sim 0)$ in the graph that we created using Pearson correlation. 
	\vspace{-1mm}
	\subsection{Method II: Hierarchical Shapley Value Explanation (H-SVE)}
	In method II, we approximate the Shapley value by a hierarchical approach: 1) detect communities in the graph, then 2) apply SVE in each community individually.
	\vspace{-3mm}
	\subsubsection{Modularity-based community detection}
	We use the same undirected graph architecture defined in \textit{Method I}, but use \textit{greedy modularity method} \cite{clauset2004finding} to divide all the features into non-overlapping communities. Then the whole features sets can be expressed by a combination of non-overlapping communities $\mathcal{N} = A_1 \bigcup A_2 \bigcup \dots \bigcup A_M$ and the features in one community only cooperate within the group, hence are independent to those in the different communities. Therefore we can define different Shapley value rules in the different communities, but the Shapley values are comparable within and across communities.
	\vspace{-3mm}
	\subsubsection{Shapley value of each feature in the community }
	With the assumption that different communities of players do not play in a game (rarely connect), we assume the communities of features are independent. In order to compare the feature importance in the whole brain, firstly we define the Shapley value for feature subset $S$ in community $A_i$ as 
	\begin{equation}
	v_{\bm{X}}(S)\coloneqq \mathbb{E}_Y[-\log\frac{1}{p(Y| \bm{X}_{A_i})}|\bm{X}_{A_i}]-\mathbb{E}_Y[-\log\frac{1}{p(Y|\bm{X}_{A_i\setminus S})}|\bm{X}_{A_i}]. 
	\end{equation}
	\textbf{Definition 2}.\textit{ Suppose the features are clustered into $\mathcal{N} = A_1 \bigcup A_2 \bigcup \dots \bigcup A_M$. The H-SVE assigns the prediction power of feature $r$ in $A_i$ by}
	\begin{equation}
	\label{hsve}
	\hat{\Phi}^H_r(v_{\bm{X}}) = \frac{1}{|A_i|}\sum_{S \subseteq A_i\setminus\{r\}} \left ( \begin{array}{c}
	|A_i|- 1 \\
	|S| 
	\end{array}
	\right ) ^ {-1}  (v_{\bm{X}}(S\cup\{r\})-v_{\bm{X}}(S)) \;.
	\end{equation}
	\textbf{Theorem 3}.\textit{ When $\bm{X}_{A_1} \perp \bm{X}_{A_2} \perp \dots \perp \bm{X}_{A_M}$, we have $	\hat{\Phi}^H_r(v_{\bm{X}}) = \Phi_r(v_{\bm{X}})$ almost surely.} 
	
    The proof is similar to the proof for \textit{Theorem 2}.
	\subsection{Monte Carlo Approximation For Large Neighborhood}
		\begin{algorithm}[t]
		\caption{Approximating the prediction power of $r$th feature's value $\Phi_r$}\label{mc}
		\hspace*{\algorithmicindent} \textbf{Input:} {$\bm{X}$, a given instance; $m$, number of samples; $v$, importance score function
			\begin{algorithmic}[1]
				\label{MC}
				\State {$\Phi_r\gets0$}
				\For{$j = 1$ to $m$}
				\State {choose a random permutation of features $\mathcal{O} \in \pi(\mathcal{N}_r)$  } 
				\State{choose a random instance $\hat{\bm{X}}$ from the training dataset }     
				\State {$v_1 \gets v(  \tau(\bm{X},\hat{\bm{X}},Pre^r(\mathcal{O})\bigcup\{r\}))$}
				\State {$v_2 \gets v(  \tau(\bm{X},\hat{\bm{X}},Pre^r(\mathcal{O})))$}    
				\State {$\Phi_r \gets \Phi_r + (v_1 - v_2)$}    
				\EndFor        
				\State{$\Phi_r \gets \frac{\Phi_r}{m}$}
			\end{algorithmic}            
			\texttt{(where $\mathcal{N}$ isthe neighborhood of $r$ in C-SVE or community of $r$ in H-SVE)}} 	
	\end{algorithm}

	Although we simplify SVE by  C-SVE or H-SVE methods, computation may still be challenging. For example: 1) in C-SVE, feature node $r$ to be analyzed is densely connected with the other nodes and 2) in H-SVE, there exists large communities. Based on the alternative formulation of the Shapley value (Eq. (\ref{origshap2})), let $\pi(\mathcal{N})$ be the set of all ordered permutations of $\mathcal{N}$. Let $Pre^r(O)$ be the set of players which are predecessors of player $r$ in the order $O\in \pi(\mathcal{N})$, we have
	\begin{equation}
		\vspace{-3mm}
	\label{origshap2}
	\Phi_r(v_{\bm{X}}) = \frac{1}{|\mathcal{N}|!}\sum_{O \in \pi(\mathcal{N})}(v_{\bm{X}}(Pre^r(O) \cup \{r\}) - v_{\bm{X}}(Pre^r(O))). 
	\end{equation}
	We use the following Monte Carlo (MC) algorithm to approximate equation (\ref{csve}) and (\ref{hsve}). We define:
		\vspace{-2mm}
	\begin{equation}
	\tau(x,\hat{x},S) = (z_1,z_2,\dots,z_s), \; \; z_i = \left \{ \begin{array}{lr}
	x_i; \; i\in S \\
	\hat{x}_i; \; i\notin S 
	\end{array}.
	\right.
		\vspace{-2mm}
	\end{equation}
	Then the unbiased MC approximation can be expressed as in Algorithm 1. Given $m$, if $2^{|\mathcal{N}(r)|} \gg m$, we will apply MC approximation.

	\section{Experiments and Results}
	\subsection{Validation on MNIST Dataset}
	In order to show the feasibility of the proposed two approaches, we test the explanation results on MNIST dataset \cite{lecun1998gradient}, where we can compare to human judgment about the feature importance. We trained a convolutional network (Conv2D(32) $\to$ Conv2D(64) $\to$ Dense(128) $\to$ Dense(10)) achieving $97.32\%$ accuracy. We parcellate the image into ROIs using \textit{slic} \cite{achanta2012slic} to mimic the setting of detecting saliency brain ROI for identifying ASD. Denoting the distance between the center of ROI $i$ and ROI $j$ as $d_{ij}$, we define the connection between ROI $i$ and $j$ as $a_{ij} = exp(-d_{ij}/2)$. Here we use $th = \frac{\sum_i\sum_j a_{ij}}{|\mathcal{E}|}$.	
	\begin{figure}[t]
	\vspace{-3mm}
		\centering
		\includegraphics[width=8cm]{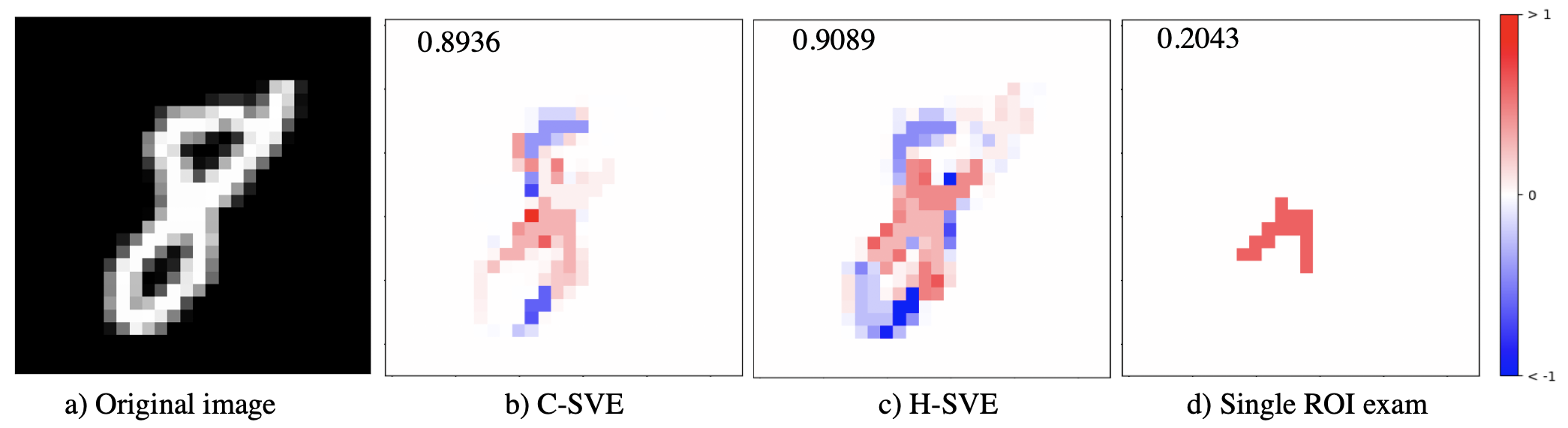}
		\caption{The predictive power for identifying (a) the digit 8 by b) C-SVE, c) H-SVE, and d) single ROI explanation. The prediction difference after corrupting the ROIs which contribute $90\%$ in total are denoted on the left corner.}
		\label{mnist_result}  
		\vspace{-3mm}
	\end{figure}
	The results are shown in Fig. \ref{mnist_result}, where we uniformly divided each ROI's importance score by the number of pixels in the ROI to mitigate dominance by large ROIs and divided by $\underset{i \in \mathcal{N}}{max}(\Phi_i)$ for visualization. The interpretation results matched our human perception that the "\textit{x cross}" shape in the center is important for recognizing digit $8$. Compared with single ROI testing, our proposed methods assigned smoother and more widely distributed importance scores to more pixels. To examine the effect of important ROIs on prediction, we corrupted pixels whose importance power added up to $90\%$ of the positive importance scores. We then compared the difference between the original prediction probability of digit 8 and the new prediction probability using the corrupted image. C-SVE and H-SVE could better fool the classifier, which decreased the prediction probability by 0.8939 and 0.9089 respectively, compared to only a 0.2043 decrease for the single ROI method. Some ROIs may not contribute to classification on their own but influence the results when combined with other regions. In the single ROI method, these ROIs will be assigned 0 importance score. However, by our proposed SVE method these ROIs can be discovered.
	
	\subsection{ASD Task-fMRI Dataset and Underlying Graph Structure}
	We tested our methods on a group of 82 children with ASD and 48 age and IQ-matched healthy controls used for training the classifiers to distinguish the two groups. Each subject underwent a biological motion perception task \cite{Kaiser07122010} fMRI scan (BOLD, TR = 2000ms, TE = 25ms, flip angle = $60^{\circ}$, voxel size $3.44\times3.44\times4 mm^3$) acquired on a Siemens MAGNETOM Trio TIM 3T scanner. We randomly split $80\%$ of the data for training, $10\%$ for validation of model parameters, and $10\%$ for testing. 
	
	The Automated Anatomical Labeling (AAL) atlas \cite{tzourio2002automated} was used to parcellate the brain into 116 regions. For each subject, we computed the $116 \times 116$ adjacency matrix using Pearson correlation. We averaged the adjacency matrix over the patient subjects in the training data and binarized the edges based on whether its weight is larger than average weight (assigning 1) or not (assigning 0). For H-SVE method, we obtained the non-overlapping community clustering for each subject by greedy modularity method \cite{newman2018networks}, which resulted in 10 communities.
	
		\vspace{-3mm}
	\subsection{Comparison with Random Forest-based Feature Importance}
\begin{figure}[t]
		\centering
		\includegraphics[width=12cm]{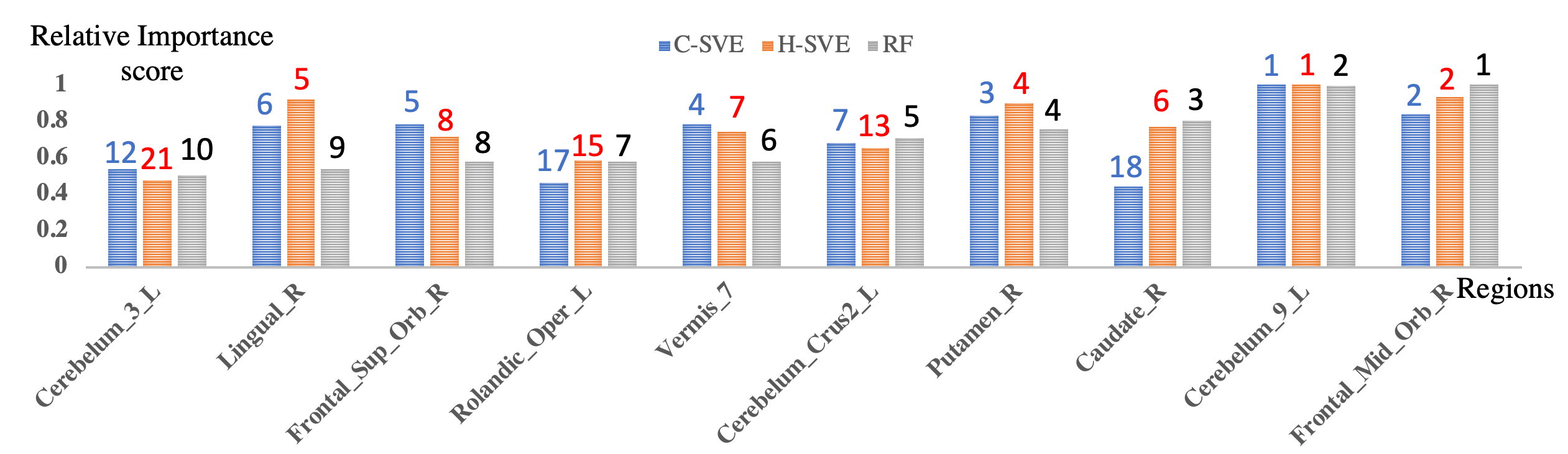}
		\caption{The relative importance scores of the top 10 important ROIs assigned by Random Forest and their corresponding importance scores in C-SVE and H-SVE. The importance rank of each ROI is denoted on the bar. }
		\label{rf}   
		\vspace{-3mm}
	\end{figure}

	As an additional "reality check" for our method, we apply a Random Forest (RF) strategy (1000 trees) to the same dataset ($71.4\%$ accuracy on testing set) and compare the results, using the RF-based feature importance (mean Gini impurity decrease) as a form of standard method for comparison. Instead of inputting the entire fMRI image, we input the node-weighted modularity, which is defined by
	$\mathcal{M}_i = \sum_{j \neq i}a_{ij}$
	where $a_{ij}$ is the partial correlation coefficient between ROI $i$ and $j$. Therefore the inputs are $116 \times 1$ vectors. Based on axiom 4, we can treat each subject as a game and each ROI as a player, and then do group-based analysis by adding $\Phi(r)$ over the subjects to investigate ROI $r$'s importance. For a fair comparison, like in RF, we used all of the training dataset. The interpretation results are shown in Fig. \ref{rf}. Seven of the top 10 important ROIs discovered by C-SVE and H-SVE overlapped with RF interpretation. 
	\vspace{-1mm}
	\subsection{Explaining The ASD Brain Biomarkers Used In Deep Convolutional Neural Network Classifier}
	\vspace{-1mm}
	\begin{figure}[b]
	\vspace{-5mm}
		\centering
		\includegraphics[width=11cm]{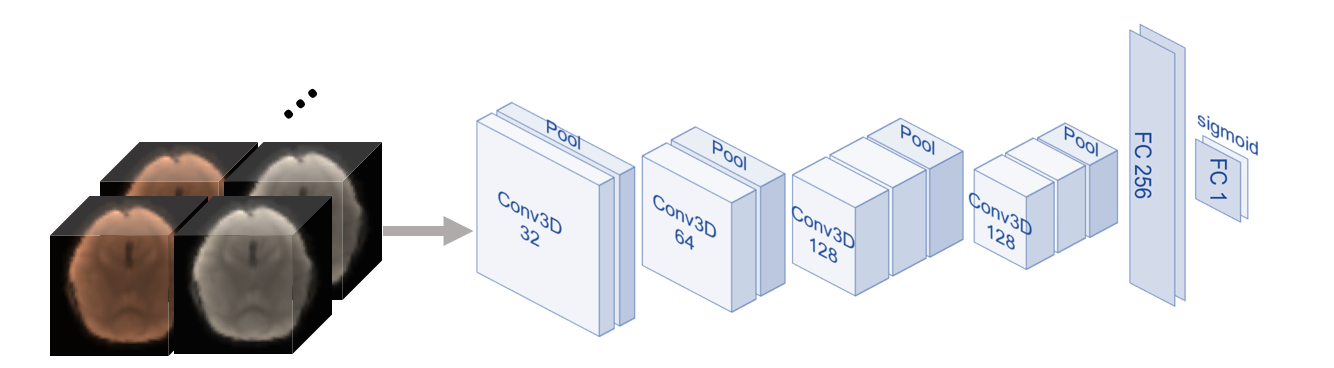}
		\caption{2CC3D network architecture }
		\label{network}  
	\end{figure}
			\begin{table}[t]
	\vspace{-2mm}
	\caption{Prediction Decrease After Corrupting Important ROIs for the DNN }
	\centering
	\begin{tabular}{p{2.5cm}<{\centering} p{2.5cm}<{\centering} p{2.5cm}<{\centering} p{2.5cm}<{\centering} }  
		\toprule
		&  C-SVE& H-SVE&  Single Region \\
		\midrule
		$\Delta prob$& 0.720 (0.221)& 0.693 (0.144) & 0.335 (0.060)  \\
		$\Delta acc$& 0.714&  0. 714 & 0.428  \\
		\bottomrule
	\end{tabular}\\
	{\footnotesize ($\Delta prob$ = decrease in test prediction probability, $\Delta acc$ = decrease in test accuracy)}
	\label{change}
	\vspace{-5mm}
\end{table}
		\begin{figure}[b]
		\centering
		\includegraphics[width=11cm]{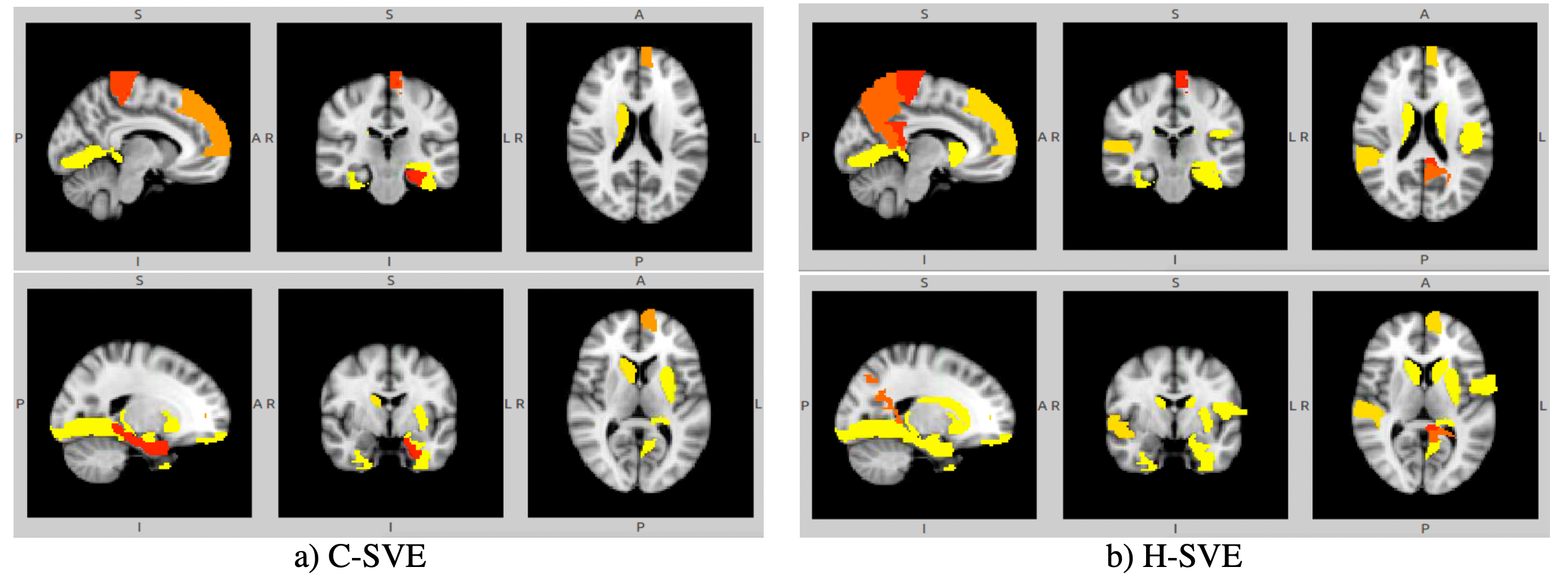}
		\vspace{-3mm}
		\caption{Top 20 predictive biomarkers detected by a) C-SVE and b) H-SVE for the deep learning classifier. More yellow ROIs signify higher importance.}
		\label{biopoint}   
	\end{figure}

	Here we chose the deep neural network 2CC3D (Fig. \ref{network}) described in \cite{li20182} using each voxel's mean and standard deviation as two channel input. We start with preprocessed 3D fMRI volumes downsampled to $32\times32\times32$. We defined the original fMRI sequence as $\bm{X}(x,y,z,t)$, the mean-channel sequence as $\bm{\tilde{X}}(x,y,z,t)$ and the standard deviation-channel as $\hat{\bm{X}}(x,y,z,t)$. For any $x,y,z$ in $\{0,1,\cdots,31\}$, $\bm{\tilde{X}}(x,y,z,t) = \frac{1}{w} \sum_{\tau=t+1-w}^{t}\bm{X}(x,y,z,\tau)$, $\,
\hat{\bm{X}}(x,\!y,\!z,\!t)^2 = \frac{1}{w-1} \sum_{\tau=t+1-w}^{t}[\bm{X}(x,\!y,\!z,\!\tau)\!-\!\bm{\tilde{X}}(x,\!y,\!z,\!t)]^2
$, where  $w$ is the temporal sliding window size and $w = 3$ in our experiment, hereby we augment data to 18720 samples. Training, validation and testing data was split based on subjects. It achieved  $85.7\%$ classification accuracy by majority voting. Running on a workstation with a Nvidia 1080 Ti GPU, testing all 7 ASD subjects in the testing dataset took $21k$ s and $26k$ s for C-SVE and H-SVE, respectively, using 1000 samples for MC approximation, which converged to the stable ranks. As in the MNIST experiment, we divided $\Phi(r)$ by the number of voxels in ROI $r$, avoiding domination by large ROIs.

\begin{figure}[t]
		\centering
		\includegraphics[width=12cm]{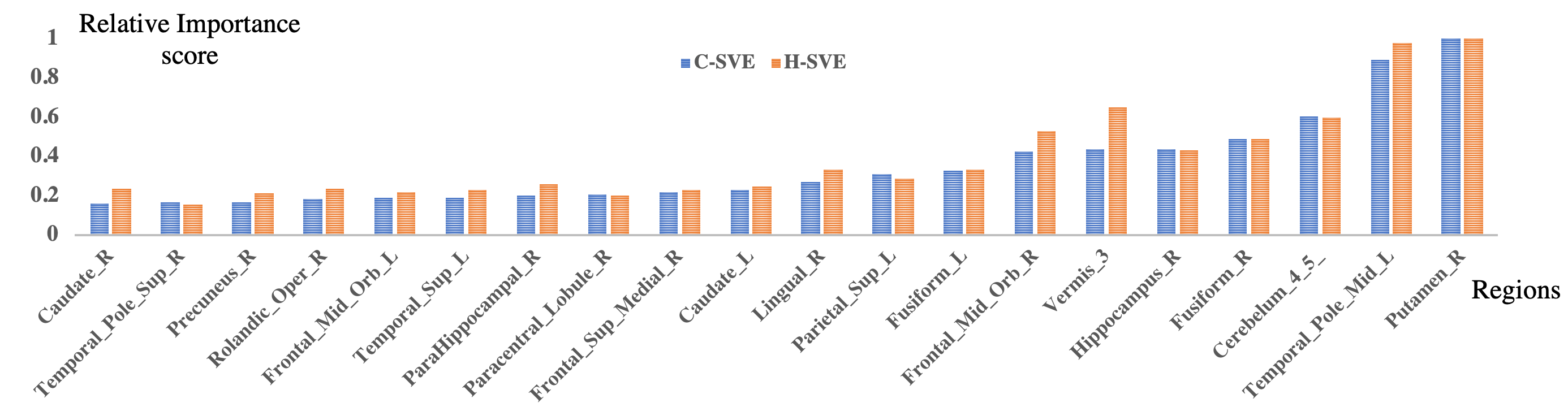}
		\vspace{-3mm}
		\caption{The relative importance scores of the top 20 ROIs assigned by C-SVE and their corresponding importance scores in H-SVE for the deep learning model.}
		\label{cnnres1}   
		\vspace{-1mm}
	\end{figure}

\begin{figure}[t]
		\centering
		\includegraphics[width=11cm]{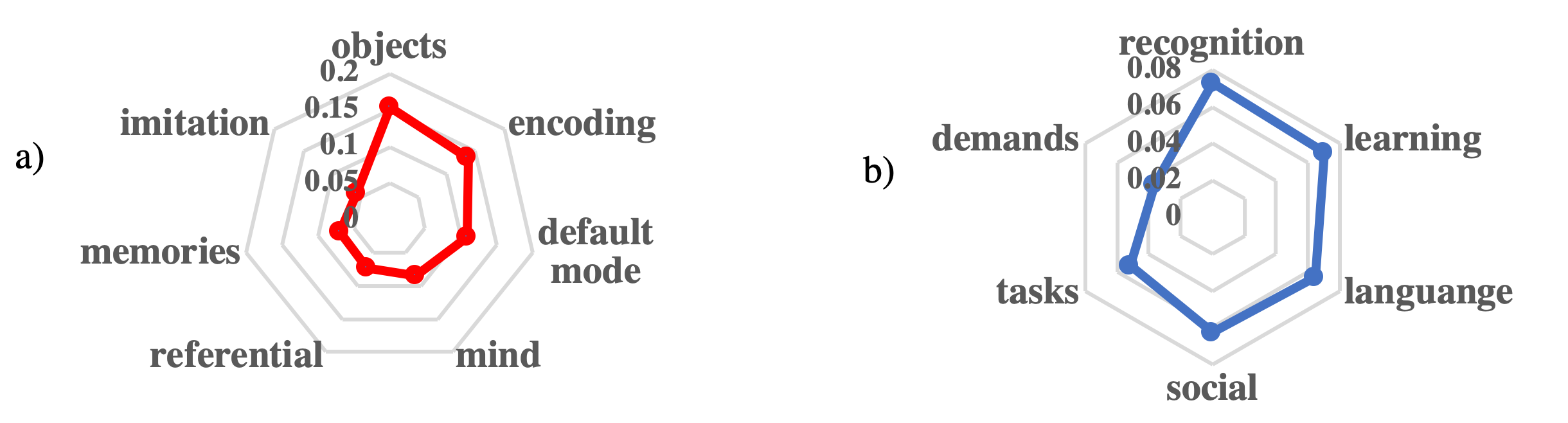}
		\vspace{-3mm}
		\caption{a) The top \textit{positive} correlations and b) the top \textit{negative} correlations between deep learning model biomarkers and functional keywords.}
		\label{cnnres2}   
		\vspace{-3mm}
	\end{figure}

The contribution/prediction power of the regions (relative to the most important one) averaged over testing subjects are illustrated in Fig. \ref{biopoint} and listed in Fig. \ref{cnnres1}. There are 19 overlapping ROIs out of the top 20 important ROIs found by C-SVE and H-SVE, although the orders were different. The \textit{Spearman rank-order correlation coefficient} \cite{young1978rating} of the importance score ranks of all the ROIs explained by both methods was $0.58$. These detected regions were consistent with the previous findings in the literature \cite{li2018brain,Kaiser07122010}.  Also, we used Neurosynth \cite{yarkoni2011large} to decode the functional keywords associated with the overlapping biomarkers found by C-SVE and H-SVE (Fig. \ref{cnnres2}). These top regions are positively related to self-referential/perspective-taking concepts (higher level social communication) and negatively related to more basic social and language concepts (lower level skills). Using the manner described in Eq. (\ref{margin}), we corrupted the important ROIs ($50\%$ of the positive importance scores summing up in order) determined by C-SVE, H-SVE, and single region testing separately and calculated the average decrease in probability $\Delta prob$ (showing mean and standard deviation) and accuracy $\Delta acc$ for the subjects in the testing set. The results are listed in Table \ref{change}.

Notice that the top 10 biomarkers we discovered using SVE in the RF model were different from the ones found in the 2CC3D model. Possible reasons are: 1) the inputs are different. 2CC3D used activation whereas RF used connectivity and 2CC3D used ASD subjects in testing set whereas RF used all the training set; 2) the prediction accuracy of RF model is much lower than 2CC3D; and 3) our proposed methods performed as a model interpreter rather than data interpreter, which may have different sensitivity response to the different models.
	
	\vspace{-3mm}
	\section{Conclusion And Future Work}
	Considering the interaction of features, we proposed two approaches (C-SVE and H-SVE) to analyze feature importance based on SVE, using the underlying graph structure of the data to simplify the calculation of Shapley value. C-SVE only considers the centralized interaction, while H-SVE uses a hierarchical approach to first cluster the feature communities, then calculate the Shapley value in each community. When a feature’s neighborhood/community still contains a large number of features, we apply MC integration method for further approximation. Experiments on the MNIST dataset showed our proposed methods can capture more interpretable features. Comparing the results with Random Forest feature interpretation on the ASD task-fMRI dataset, we further validated the accuracy and feasibility of the proposed methods. When applying both methods on a deep learning model, we discovered similar possible brain biomarkers, which matched the findings in the literature and had meaningful neurological interpretation.  The pipeline can be generalized to other feature importance analysis problems, where the underlying graph structure of features is available.  
	
	Our future work includes testing the methods on different atlases, graph building methods, and  community clustering methods, etc. In addition, the interaction score is embedded in the proposed algorithms. It can be disentangled to understand the interaction between the features.
		\appendix
	\section{Appendix: Proof of Theorem 2}
For any subset $A\subset\mathcal{N}$, we use the short notation $U_{r}(A)\coloneqq A\cap\mathcal{N}_{r}$
and $V_{r}(A)\coloneqq A\cap(\mathcal{N}\setminus\mathcal{N}_{r})$,
noting that $A=U_{r}(A)\cup V_{r}(A)$. Rewriting Eq. (\ref{origshap})
as \vspace{-2mm}
 
\[
\Phi_{r}(v_{\bm{X}})=\frac{1}{|\mathcal{N}|}\!\sum_{U\subseteq\mathcal{N}_{r}\setminus\{r\}}\!\sum_{A\subseteq\mathcal{N},U_{r}(A)=U}\!\left(\begin{array}{c}
|\mathcal{N}|\!-\!1\\
|A|
\end{array}\right)^{-1}\!(v_{\bm{X}}(A\cup\{r\})\!-\!v_{\bm{X}}(A)),
\]
\vspace{-2mm}
and using \vspace{-2mm}
 
\[
\sum_{A\subseteq\mathcal{N},U_{r}(A)=U}\left(\begin{array}{c}
|\mathcal{N}|-1\\
|A|
\end{array}\right)^{-1}=\frac{|\mathcal{N}|}{|\mathcal{N}_{r}|}\left(\begin{array}{c}
|\mathcal{N}_{r}|-1\\
|U|-1
\end{array}\right)^{-1},
\]
\vspace{-3mm}
the expected error between $\hat{\Phi}_{r}^{C}(v_{\bm{X}})$ and $\Phi_{r}(v_{\bm{X}})$
is 
\[
\mathbb{E}[\vert\hat{\Phi}_{r}^{C}(v_{\bm{X}})\!-\!\Phi_{r}(v_{\bm{X}})\vert]\!\leq\!\frac{1}{|\mathcal{N}|}\!\sum_{U\subseteq\mathcal{N}_{r}\setminus\{r\}}\!\sum_{A\subseteq\mathcal{N},U_{r}(A)=U}\left(\begin{array}{c}
|\mathcal{N}|\!-\!1\\
|A|
\end{array}\right)^{-1}\!\mathbb{E}[\vert\Delta_{r}^{{\bm{X}}}(U,A)\vert]
\]
where 
\begin{align*}
\Delta_{r}^{{\bm{X}}}(U,A) & =(v_{\bm{X}}(U\cup\{r\})-v_{\bm{X}}(U))-(v_{\bm{X}}(A\cup\{r\})-v_{\bm{X}}(A))\\
 & =\log\frac{p(Y\vert X_{\mathcal{N}\setminus U})}{p(Y\vert X_{\mathcal{N}\setminus(U\cup\{r\})})}-\log\frac{p(Y\vert X_{\mathcal{N}\setminus(U\cup V)})}{p(Y\vert X_{\mathcal{N}\setminus(U\cup V\cup\{r\}})},
\end{align*}
with $V$ short for $V_{r}(A)$. Let $W=\mathcal{N}\setminus(\mathcal{N}_{r}\cup V)$,
$Z=\mathcal{N}_{r}\setminus(\{r\}\cup U)$. Then 
\begin{align}
\Delta_{r}^{{\bm{X}}}(U,A) & =\log\frac{p(Y\vert X_{W\cup V\cup Z\cup\{r\}})p(Y\vert X_{W\cup Z})}{p(Y\vert X_{W\cup V\cup Z})p(Y\vert X_{W\cup Z\cup\{r\}})}.\label{deltar}
\end{align}
Since $X_{r}\perp X_{V}\vert X_{Z}$, we have $p(X_{V}\vert X_{W\cup Z\cup\{r\}})=p(X_{V}\vert X_{W\cup Z})$,
and 
\[
(\star)=\frac{p(X_{W\cup V\cup Z\cup\{r\}})p(X_{W\cup Z})}{p(X_{W\cup V\cup Z})p(X_{W\cup Z\cup\{r\}})}\!=\!\frac{p(X_{W\cup Z\cup\{r\}})p(X_{V}\vert X_{W\cup Z\cup\{r\}})p(X_{W\cup Z})}{p(X_{W\cup Z})p(X_{V}\vert X_{W\cup Z})p(X_{W\cup Z\cup\{r\}})}=1.
\]
We can multiply the quotient in Eq. (\ref{deltar}) by $(\star)$,
\begin{align*}
\Delta_{r}^{{\bm{X}}}(U,A) & =\log\frac{p(Y\vert X_{W\cup V\cup Z\cup\{r\}})p(Y\vert X_{W\cup Z})}{p(Y\vert X_{W\cup V\cup Z})p(Y\vert X_{W\cup Z\cup\{r\}})}\frac{p(X_{W\cup V\cup Z\cup\{r\}})p(X_{W\cup Z})}{p(X_{W\cup V\cup Z})p(X_{W\cup Z\cup\{r\}})}\\
 & =\log\frac{p(X_{W\cup V\cup\{r\}}\vert Y,X_{Z})p(Y,X_{Z})p(Y,X_{Z})p(X_{W}\vert Y,X_{Z})}{p(Y,X_{Z})p(X_{W\cup V}\vert Y,X_{Z})p(Y,X_{Z})p(X_{W\cup\{r\}}\vert Y,X_{Z})}.
\end{align*}
We have $p(X_{W\cup V\cup\{r\}}\vert Y,X_{Z})=p(X_{W\cup V}\vert Y,X_{Z})p(X_{r}\vert Y,X_{Z})$,
since $X_{W\cup V}\perp X_{r}\vert Y,X_{Z}$. So 
\[
\Delta_{r}^{{\bm{X}}}(U,A)=\log\frac{p(X_{W\cup V}\vert Y,X_{Z})p(X_{r}\vert Y,X_{Z})p(X_{W}\vert Y,X_{Z})}{p(X_{W\cup V}\vert Y,X_{Z})p(X_{W\cup\{r\}}\vert Y,X_{Z})}.
\]
Since $X_{W}\perp X_{r}\vert Y,X_{Z}$, we have $p(X_{W\cup\{r\}}\vert Y,X_{Z})=p(X_{W}\vert Y,X_{Z})p(X_{r}\vert Y,X_{Z})$.
Hence $\Delta_{r}^{{\bm{X}}}(U,A)=\log1=0$. Therefore we have $\mathbb{E}[\vert\hat{\Phi}_{r}^{C}(v_{\bm{X}})-\Phi_{r}(v_{\bm{X}})\vert]=0$
.
	\medskip
	\vspace{-3mm}
	\small
	\bibliographystyle{ieeetr}
	\bibliography{biomarker}
\end{document}